%% file: main.tex
\documentclass[runningheads]{llncs}

\usepackage[T1]{fontenc}
\usepackage{graphicx}
\usepackage{booktabs}
\usepackage{amsmath}
\usepackage{amssymb}
\usepackage[table]{xcolor}
\usepackage{multirow}
\usepackage{array}
\usepackage{subcaption}
\usepackage{enumitem}
\usepackage{placeins}
\usepackage{float}
\usepackage{makecell}
\usepackage{xspace}

\input{_macros}

\usepackage{hyperref}
\hypersetup{colorlinks=true, linkcolor=blue, citecolor=blue, urlcolor=blue}

\usepackage[capitalize]{cleveref}
\crefname{section}{Sec.}{Secs.}
\crefname{table}{Table}{Tables}
\crefname{figure}{Fig.}{Figs.}

\begin{document}


\title{SAFE: Scene-Aware Feature Modulation for Color Constancy with Learned Color Space in Pure-Color Scenes}
\titlerunning{SAFE with LCS}

\author{Yuan-Kang Lee\inst{1} \and Kuan-Lin Chen\inst{2} \and Chih-Heng Chang\inst{2} \and Jian-Jiun Ding\inst{2}}
\authorrunning{Y.-K.~Lee et al.}
\institute{\textsuperscript{\rm 1} MediaTek Inc., Hsinchu, Taiwan \\ \textsuperscript{\rm 2} National Taiwan University, Taipei, Taiwan\\
}

\maketitle
\input{teaser}


\input{00_abstract}
\input{01_intro}
\input{02_related}
\input{04_method}
\input{05_experiment}
\input{10_conclusion}

\FloatBarrier
\bibliographystyle{splncs04}
\bibliography{references}

\end{document}

%% file: _macros.tex
\newcolumntype{C}{>{\centering\arraybackslash}p{1.05cm}}

\definecolor{lightred}{RGB}{253,191,191}
\definecolor{lightorange}{RGB}{255,223,191}
\definecolor{lightyellow}{RGB}{254,240,198}

%% file: teaser.tex
\begin{center}
    \centering
    \includegraphics[width=\linewidth, trim={0cm 0cm 0cm 0cm},clip]{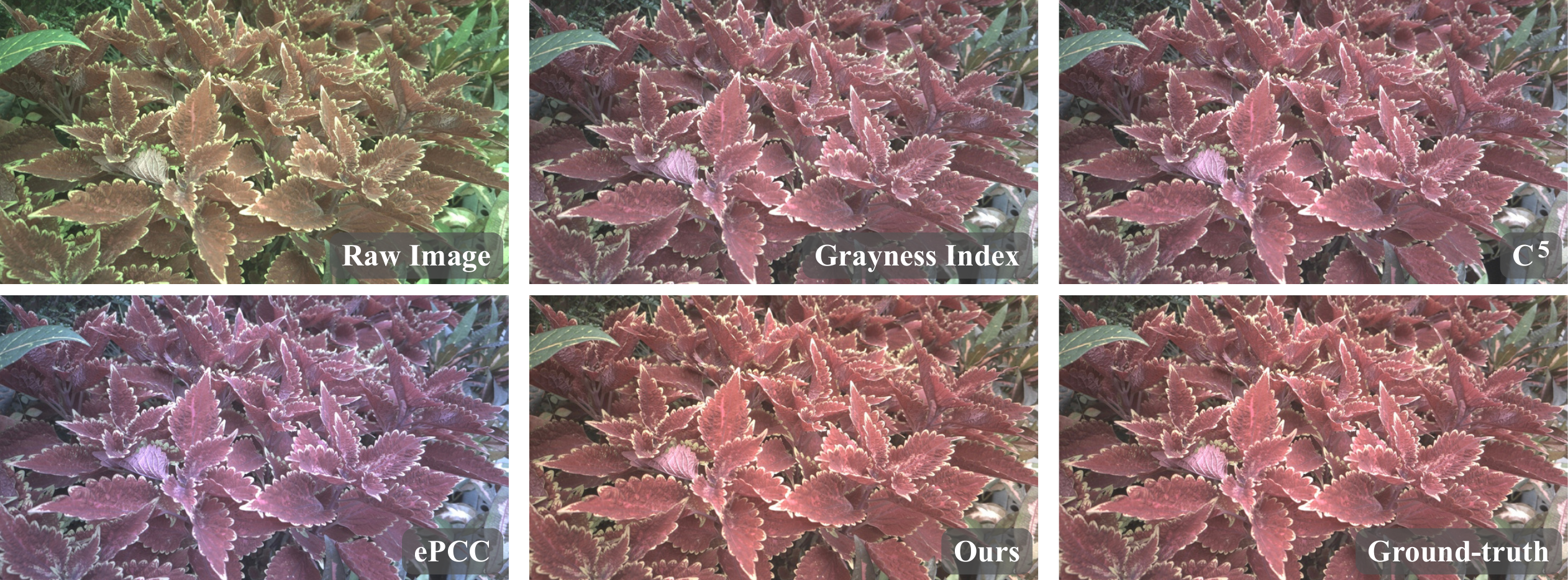}
    \captionof{figure}{
        \textbf{Comparison on pure-color scenes.} 
        Images are gamma-corrected for visualization. Our approach effectively eliminates color casts in pure color scenes, outperforming both supervised and unsupervised state-of-the-art methods.
    }
    \label{fig:teaser}
\end{center}

%% file: 00_abstract.tex
\begin{abstract}

Color constancy on pure-color scenes is
challenging: when most pixels share a narrow band of hues, every chromaticity-based cue collapses to a single point and standard estimators become ambiguous. We propose a compact framework that couples two innovations: (i) SAFE, a Scene-Aware FeaturE modulation network that organizes illumination cues into a structured four-token representation, which is then selectively reweighted based on scene complexity features; (ii) the Learned Color Space (LCS), a scene-dependent chromaticity normalization that directly addresses the chromaticity collapse problem for pure-color scenes. Experiment results show that SAFE consistently improves performance in pure-color scenes. Compared to the best-performing baseline in each metric, it reduces the mean angular error by 10\%, the best-25\% error by 20\%, and the worst-25\% error by 5.8\%.

\keywords{Color constancy \and Pure color \and Illuminant estimation}

\end{abstract}


%% file: 01_intro.tex
\section{Introduction}
\label{sec:intro}

The Image Signal Processor (ISP) is a crucial component in modern imaging systems, comprising various processing modules to convert sensor RAW data into sRGB space~\cite{mohammadi2026low,zhang2023mm,tian2023multi,zhang2022idr,lee2024efficient,liang2022semantically,hashemi2010image,kim2023paramisp,rim2022realistic,jiao2026efficient}. Among these, color constancy plays the most vital role, aiming to estimate the illuminant of the scene to achieve color-invariant image representation. Traditional color constancy research has been mainly devoted to the well-illuminated single-light-source scenario, and existing methods have already demonstrated impressive performance. However, in real-world scenarios, conditions are often far from ideal and state-of-the-art research is exploring the more challenging frontiers. These include: (1) Multi-illuminant color constancy, which requires estimating a pixel-wise illumination map to account for spatial lighting distributions ~\cite{kim2021large,kim2024attentive,entok2024pixel,luo2026enhancing}; (2) Nighttime color constancy, aimed at extracting accurate illuminant information under extreme low-light and high-noise conditions ~\cite{cheng2024nighttime,li2024nightcc,Lee2026RLAWB}; and (3) sRGB white balance, which seeks to restore true scene colors within the non-linear and physically inconsistent sRGB domain ~\cite{afifi2019color,afifi2020deep,farghaly2023two,li2023swbnet,serrano2025revisiting,cheng2026perception}. 

In addition to these difficulties, there is another, more common but more severe problem that occurs when scene color distributions become too uniform, so that scene reflectance is indistinguishable from illuminant color. Such situations are ubiquitous in macro photography, in close-up portraits, in vast landscapes (e.g., oceans, skies, or snow), and in stage lighting at concerts. In these pure-color scenes, the image formation process is inherently ambiguous: a white illuminant on a red surface produces the same observation as a red illuminant on a white surface. Classical statistics-based methods rely on global assumptions that collapse in these constrained scenarios \cite{buchsbaum1980spatial,finlayson2004shades}. Simultaneously, most learning-based methods \cite{yu2020cascading,afifi2020deep,afifi2021cross,afifi2022auto,tang2022transfer,kim2025ccmnet} are trained on general datasets where the diverse context is plentiful, leading to biased estimations, as demonstrated in Fig.~\ref{fig:teaser}.

To address these limitations, we introduce \emph{SAFE}, a scene-aware feature modulation network whose scene complexity modulation selectively reweights a structured four-token illumination representation, and the \emph{Learned Color Space} (LCS), a scene-dependent chromaticity normalization that directly counters the chromaticity
collapse to which standard normalizations succumb on pure-color scenes. Our design rests on two parallel observations: first, different color cues offer varying degrees of informativeness across diverse scenes; second, the uniform-channel chromaticity normalization that underpins standard illumination features becomes precisely adversarial when the scene's chromaticity distribution collapses. Fig.~\ref{fig:flowchart} illustrates the overall flowchart of our proposed method. The contributions are summarized as follows:

\begin{itemize}

    \item A \textbf{four-token illumination feature design} that
    combines physics-derived cues (specular log-chromaticity,
    dichromatic principal direction) with chromaticity statistics from original and edge images into a structured 24D representation with clear semantic roles.

    \item The \textbf{Scene-Aware Feature Modulation} (SAFE) network that adaptively reweights illumination tokens based on scene characteristics, enabling robust and stable behavior across diverse imaging conditions.

    \item The \textbf{Learned Color Space} (LCS): a scene-dependent chromaticity normalization predicted by a lightweight model from the image and scene complexity features, designed to counter chromaticity collapse on pure-color scenes.

\end{itemize}

\input{Flowchart}

%% file: Flowchart.tex
\begin{figure*}[!t]
    \centering
    \includegraphics[width=\linewidth, trim={0cm 0cm 0cm 0cm},clip]{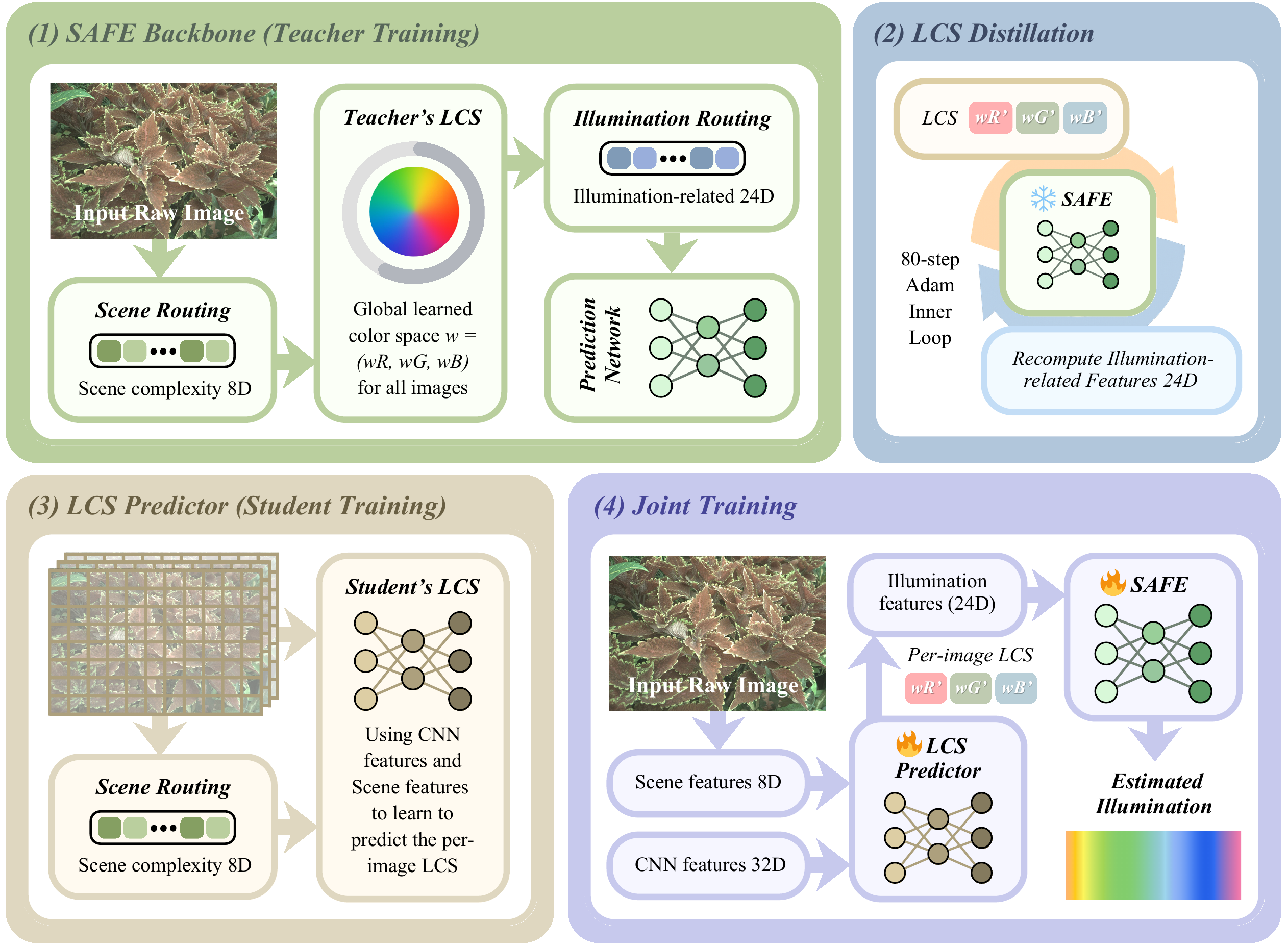}
\vspace{-5mm}
    \caption{\textbf{Overview of the proposed color constancy framework.} 
    Training proceeds in four phases: \textbf{(1)} pre-train the SAFE backbone $\phi$ alongside a coarse global $w$-module under the angular loss; \textbf{(2)} freeze $\phi^{(1)}$ and search for a scene-dependent color axis that minimizes the teacher's angular error; \textbf{(3)} supervise the LCS predictor $\psi$ on the oracle
    in logit space; \textbf{(4)} jointly fine-tune $\phi$ and $\psi$
    end-to-end. At inference, the LCS predictor $\psi$ outputs a
    per-image color axis $\mathbf{w}(\mathbf{I})$ that re-parameterizes
    the chromaticity normalization for calculating the four-token illumination
    feature $\mathbf{f}_w$; the SAFE backbone then modulates $\mathbf{f}_w$
    via a scene descriptor to predict the
    illuminant estimate $\hat{\mathbf{y}}$.}
    \label{fig:flowchart}
\end{figure*}


%% file: 02_related.tex
\section{Related Work}
\label{sec:related}

\textbf{Statistical Color Constancy.} Statistical methods exploit scene statistics via achromatic averages \cite{buchsbaum1980spatial}, local space averages ~\cite{ebner2004parallel,song2019neural,ulucan2023block,ulucan2024computational}, color histogram coincidence ~\cite{jiang2012flexible,jiang2012auto,zhang2016automatic}, maximum responses \cite{land1977retinex}, and their unification via Minkowski norms \cite{finlayson2004shades}. Edge-based methods \cite{van2007edge} extend these priors to image derivatives with photometric weighting \cite{gijsenij2010generalized}. Other statistical methods estimate the illuminant by leveraging the achromaticity of pixels ~\cite{Yang2015Efficient,qian2019finding,cheng2024nighttime}. While computationally efficient, these  heuristic methods fundamentally assume high chromatic diversity. In pure color scenes, these assumptions may be violated. 

\vspace{1pt} 

\noindent \textbf{Learning-Based Color Constancy.} Early learning-based methods use CNNs with confidence-weighted pooling \cite{hu2017fc4} and log-chrominance localization \cite{barron2015convolutional}. Recent progress applies representation learning via contrastive frameworks \cite{lo2021clcc} for feature discrimination and diffusion-based models \cite{chang2025gcc} for learning sensor-agnostic illumination priors. A notable line of work is based on Fourier-domain illuminant estimation, reformulating global illuminant estimation as a spatial localization problem on the UV histogram representation~\cite{barron2017fast,wei2025integral}. Another important research direction focuses on cross-camera color constancy, where models must generalize to unseen camera sensors with different spectral sensitivities. $C^5$~\cite{afifi2021cross} adopts a transductive inference strategy by leveraging unlabeled test-time images to calibrate the model to the target camera’s spectral characteristics; CCMNet~\cite{kim2025ccmnet} leverages calibrated color correction matrices commonly available in camera ISPs to construct compact camera fingerprint embeddings.  In parallel, reinforcement learning has been explored to optimize the parameters of the color constancy algorithm via sequential decision-making \cite{Lee2026RLAWB}. However, most learning-based methods struggle in low-diversity scenarios where semantic cues are absent.

\vspace{1pt}

\noindent \textbf{Color Constancy for Pure Color Scenes.} In recent research \cite{Wei2023ColorCF,Liu2026Color}, chromaticity features are adopted as the key indicators for estimating illuminant in pure color environments. Nonetheless, existing methods often fail to account for scene complexity. Instead of discerning specific scene characteristics, they tend to apply learned statistical patterns indiscriminately across diverse environments. Feature modulation techniques~\cite{Perez2018FiLM,Wang2018DeepSFT,Sun2023SAFM} have been demonstrated to be effective in conditioning neural networks on auxiliary information. Unlike approaches designed for discrete categorical priors, pure color constancy requires modulation based on a confidence measure derived from chromatic distribution analysis. Our method improves the performance of feature usage in predicting the illuminant using the designed feature modulation mechanism.

\vspace{1pt}

\noindent \textbf{Color Space Design in White Balance Correction.} Cheng \textit{et al.} introduced a novel white balance correction framework based on a Learnable HSI (LHSI) color space~\cite{cheng2026perception}. By introducing learnable parameters, this framework enhances the flexibility of color representation and further improves the performance of white balance. Drawing inspiration from this design, we embed Learned Color Space (LCS) into our pure color constancy model. This enables the model to tailor the color space representation according to scene-specific characteristics, thereby enhancing the precision of illuminant estimation.  

%% file: 04_method.tex
\section{Proposed Method}
\label{sec:method}

The proposed framework consists of three components: (1) a feature extractor to separate scene-complexity cues from illumination-related statistics, (2) a feature modulation network to treat scene complexity as a selective gate for illumination features, prior to the predictor, and (3) a scene-dependent LCS that adaptively re-parameterizes per-image chromaticity to achieve better illuminant estimation in pure-color scenes. 

\subsection{Feature Extraction}

We extract two families of features: scene features that summarize scene color complexity and illumination-related features that expose the illuminant chromaticity through physical and statistical cues. Throughout, for a pixel with RGB values $(R,G,B)$ in an image $\textbf{I}$, the chromaticity $(r,g)$ is defined as: 

\vspace{-2pt}

\begin{equation}
{(r,g)} = \left( \frac{R}{R+G+B}, \frac{G}{R+G+B} \right).
\end{equation}

\noindent \textbf{Scene Complexity Features.} We compute a single 8D scene descriptor shown in Eq. (2) that characterizes the complexity of the scene's color distribution to enable adaptive feature weighting.

\vspace{-2pt}

\begin{equation}
\boldsymbol\rho(\mathbf{I}) = \bigl[\,
\underbrace{H,\;\pi,\;\sigma_r,\;\sigma_g}_{\\\text{color-blind subset}},\;\;
\underbrace{\bar r,\;\bar g,\;r_{\max},\;g_{\max}}_{\text{color-aware extension}}
\,\bigr]\;\in\mathbb{R}^{8}.
\end{equation}

\noindent The scene descriptor is used in two distinct ways in the rest of the network:the color-blind subset $\mathbf{g}\equiv\boldsymbol\rho{[0:4]}\in\mathbb{R}^{4}$ drives the modulation gate, while the full descriptor
$\boldsymbol\rho$ drives the Learned Color Space (LCS) predictor. The two subnetworks
require different views of the scene, hence the design. Each feature in the scene descriptor is introduced as follows:

\begin{itemize}[leftmargin=*,noitemsep,topsep=0pt]

\item \textit{Chromaticity Entropy \textit{H}.} We construct a 2D histogram over the chromaticity space using $32 \times 32$ bins and compute the normalized Shannon entropy:

\vspace{-1pt}

\begin{equation}
H = -\frac{1}{\log(N_{\text{bins}}^2)} \sum_{i,j} p_{ij} \log(p_{ij} + \epsilon),
\end{equation}

\vspace{-1pt}

where $N_{\text{bins}}$ is the number of bins per dimension (set to 32), $p_{ij}$ is the probability of chromaticity falling in bin $(i,j)$ and $\epsilon$ is a small constant for stability.  

\item \textit{PCA Variance Ratio $\pi$.} We extract the primary explained variance ratio from the 2D chromaticity distribution via principal component analysis.

\vspace{1pt}

\item \textit{Chromaticity dispersion components $\sigma_r$, $\sigma_g$.} We measure the standard deviation of the chromaticity distribution to supply the absolute scale of dispersion.

\vspace{1pt}

\item \textit{Chromaticity of RGB mean $\bar r,\;\bar g$ and max $r_{\max},\;g_{\max}$.} We compute these four features to expose the scene's dominant chromaticity.

\end{itemize}

\noindent The 8D features describe the same chromaticity distribution along two orthogonal axes: shape and location. The color-blind subset is invariant under translation in chromaticity $(r,g)$ space. They describe how the chromaticity content is spread without saying which hue dominates. The color-aware subset explicitly encodes where in the simplex the distribution sits, i.e., the dominant hue.


\vspace{1pt}

\noindent \textbf{Illumination-Related Features.} These proposed 24D illumination-related features are organized as four semantic tokens:

\begin{itemize}

\item \textit{Global chromaticity token $\mathbf{A}$}. The chromaticity coordinates of four representative pixels from the original image are shown in Eq. (4).

\vspace{-2pt}

\begin{equation}
\mathbf{A} = \bigl[\;
{(r_{},g_{})_{\max}},\;\;
{(r_{},g_{})_{\text{mean}}}_{\text{}},\;\;
{(r_{},g_{})_{\text{bright}}}_{\text{}},\;\;
{(r_{},g_{})_{\text{dark}}}_{\text{}} \; \bigr].
\end{equation}

\begin{enumerate}

\item Max chromaticity: the maximum value in each color channel.

\item Mean chromaticity: the average value in each color channel.

\item Brightest chromaticity: the chromaticities of the brightest pixels, determined by the highest $(R+G+B)$ sum within a valid intensity range. We exclude pixels whose intensity exceeds 98\% of the maximum intensity. 

\item Darkest chromaticity: the chromaticities of the darkest pixel above the noise floor, identified by the lowest $(R+G+B)$ sum while filtering out the underexposed. We exclude pixels below 2\% of the maximum intensity.

\end{enumerate}

\noindent Our method mitigates sensitivity to sensor saturation and noise by averaging the chromaticities of the top-k brightest and darkest pixels. We set k=20 for our experimental configuration.

\item \textit{Local edge chromaticity token $\mathbf{B}$.} Complementing the original image, the edge map offers alternative insights into the illumination of the scene. Denoting the chromaticity of the Sobel edge image as $r_e$ and $g_e$, we follow the same procedure to extract the four representative features for token $\mathbf{B}$.

\vspace{-2pt}

\begin{equation}
\mathbf{B} = \bigl[\;
{(r_{e},g_{e})_{\max}},\;\;
{(r_{e},g_{e})_{\text{mean}}}_{\text{}},\;\;
{(r_{e},g_{e})_{\text{bright}}}_{\text{}},\;\;
{(r_{e},g_{e})_{\text{dark}}}_{\text{}} \; \bigr].
\end{equation}

Edges emphasize inter-region transitions and expose illumination-related information hidden in 
smooth-region averages. 

\item \textit{Specular and dispersion token $\textbf{C}$}. We use a dual-criterion filtering process to identify candidate specular pixels. First, we identify high-intensity regions where pixel values exceed 70\% of the image's global maximum. Second, we apply a low-saturation constraint, requiring the normalized difference between the maximum and minimum color channels to remain below 0.2. The low saturation criterion distinguishes specular highlights from bright but saturated material colors. We then compute log-chromaticity ratios as shown in the first two dimensions of Eq. (6), where a small constant $\epsilon$ is added to the denominator to avoid undefined values when $G=0$. The remaining four dimensions describe how spread-out the chromaticity is both on the original image and on the Sobel edge image. Together, token \textbf{C} combines a direct physical illumination cue with two statistical concentration measures.

\vspace{-5pt}

\begin{equation}
\mathbf{C} = \bigl[\,
\underbrace{ \left( \log\left(\frac{R}{G+\epsilon}\right),\ \log\left(\frac{B}{G+\epsilon}\right) \right) }_{\text{specular log-chrom.}\text{}},\;\;
\underbrace{(\sigma_{r},\sigma_{g})_{\text{orig}}}_{\text{global chrom.}},\;\;
\underbrace{(\sigma_{r},\sigma_{g})_{\text{edge}}}_{\text{edge chrom.}}
\,\bigr].
\end{equation}



\vspace{1pt}

\item \textit{Dichromatic geometry token \textbf{D}.} The geometry descriptor is a physics-derived illuminant cue grounded in the Dichromatic Reflection Model ~\cite{shafer1985using,tan2003illumination,tan2004color,zhao2022improving}. We extract this geometry as the first principal direction of squared-brightness weighted chromaticity as shown in Eq. (7):

\begin{equation}
\mathbf{D} = \mathrm{argmax}_{}\;\mathbf{u}^\top \!\Bigl( \ \!\!\sum_{p\in\mathcal{V}}
\,(R_p\!+\!G_p\!+\!B_p)^2\;
\bigl(\mathbf{c}(p)-\bar{\mathbf{c}}\bigr)\bigl(\mathbf{c}(p)-\bar{\mathbf{c}}\bigr)^\top\!\!\ \Bigr)\mathbf{u},
\end{equation}
\vspace{1pt}

\noindent where $p \in \mathcal{V}$ indexes a valid pixel, $\mathbf{c}(p) = (r(p), g(p))$ is its chromaticity, $\bar{\mathbf{c}}$ is the squared-brightness weighted mean of $\{\mathbf{c}(p)\}$, and $\mathbf{u} \in \mathbb{R}^{2}$ is a unit vector in the chromaticity plane. This direction aligns with the line spanned by surface chromaticity and illuminant
chromaticity in $r$-$g$ space, exposing the illuminant axis modulo a
1D ambiguity. The sign of the eigenvector is fixed by requiring its
first component to be non-negative.

\end{itemize}

\subsection{Feature Modulation Network} 
The feature modulation network mixes 
\textbf{g} (how complex the scene is) with $\mathbf{f_{illum}}=[\ \mathbf{A};\mathbf{B};\mathbf{C};\mathbf{D} \ ]$ (what colors are present) to estimate illumination. 

\vspace{1pt}

\noindent \textbf{Scene-conditioned attention.} The scene complexity vector $\mathbf{g}\in\mathbb{R}^{4}$ produces a 24D
multiplicative attention mask via a two-layer MLP with a sigmoid output:

\vspace{-5pt}

\begin{equation}
\boldsymbol\alpha(\mathbf{g}) = \sigma\! \, \bigl(\,W_2 \,\mathrm{GELU}(W_1\,\mathbf{g} + \mathbf{b}_1) + \mathbf{b}_2\,\bigr)\;\in[0,1]^{24},
\end{equation}

\vspace{2pt}

\noindent with $W_1 \in \mathbb{R}^{d_g \times 4}$ and $W_2 \in \mathbb{R}^{24 \times d_g}$ the
linear weights, $\mathbf{b}_1 \in \mathbb{R}^{d_g}$ and $\mathbf{b}_2 \in \mathbb{R}^{24}$
the corresponding bias vectors, and $d_g$ the hidden width. $\mathrm{GELU}(\cdot)$ denotes the Gaussian Error Linear Unit
activation~\cite{hendrycks2016gaussian} applied element-wise, and $\sigma(\cdot)$ is
the element-wise sigmoid function. The mask is then applied
element-wise to $\mathbf{f_{illum}}$ to produce the modulated illumination features $\mathbf{f_{m}}$:

\vspace{-4pt}

\begin{equation}
{\mathbf{f_{m}}} \;=\; \boldsymbol\alpha(\mathbf{g})\,\odot\,\mathbf{f_{illum}} \;\in\mathbb{R}^{24}.
\end{equation}


\noindent \textbf{Fusion and Backbone.} The modulated illumination features ${\mathbf{f}}_m$
are concatenated with the scene complexity vector $\mathbf{g}$ and fed into
a residual-MLP backbone $\phi$ that consists of an input projection, $L$
residual blocks, and a linear output head. The internal hidden states are
denoted $\mathbf{h}_0,\dots,\mathbf{h}_L \in \mathbb{R}^{d}$. The 28D concatenated input is linearly projected to
the $d$-dimensional working space, passed through a GELU, and
LayerNorm-normalised to form the initial hidden state:

\vspace{-3pt}

\begin{equation}
\mathbf{h}_0 = \mathrm{LN} \, \!\bigl(\,\mathrm{GELU}(\, W_0\,[{\mathbf{f}}_m;\,\mathbf{g}] + \mathbf{b}_0 \, )\,\bigr) \;\in\mathbb{R}^{d}.
\end{equation}

\noindent For $\ell = 1,\dots,L$, each block transforms the
previous hidden state via a linear layer, GELU, LayerNorm and dropout,
and adds the result back residually:

\vspace{-3pt}

\begin{equation}
\mathbf{h}_\ell = \mathbf{h}_{\ell-1} + \mathrm{Drop} \, \!\bigl(\,\mathrm{LN}( \, \mathrm{GELU}(\,W_\ell\,\mathbf{h}_{\ell-1} + \mathbf{b}_\ell\,))\,\bigr) \;\in\mathbb{R}^{d}.
\end{equation}

\noindent The skip connection lets each block refine $\mathbf{h}_{\ell-1}$, stabilizing the optimization of the $L$-block stack and
letting gradients propagate directly back to the input projection.
$\mathrm{Drop}(\cdot)$ is element-wise dropout applied only during training. Finally, a single linear layer maps the final hidden state to the 3D illuminant prediction:

\vspace{-3pt}

\begin{equation}
\hat{\mathbf{y}} = W_{\text{out}}\,\mathbf{h}_L \;\in\mathbb{R}^{3}.
\end{equation}

\subsection{Learned Color Space} 

The components above all depend on the chromaticity normalization
$(r,g) = (R,G)/(R+G+B)$. It implicitly assumes that the
three channels should be weighted equally. However, this assumption fails in chromaticity-deficient scenes where most pixels share a single hue. The proposed Learned Color Space (LCS) addresses this limitation by introducing image-specific learnable channel weights.

Let $\mathbf{w} = (w_R, w_G, w_B)$ live on the probability
simplex. The LCS replaces the uniform normalization
$1/(R+G+B)$ with the $w$-weighted normalization
$1/({w_R R + w_G G + w_B B})$, producing the generalized
chromaticity:

\vspace{-0.3cm}

\begin{equation}
r_w(p) = \frac{w_R R_p}{w_R R_p + w_G G_p + w_B B_p},\quad
g_w(p) = \frac{w_G G_p}{w_R R_p + w_G G_p + w_B B_p}.
\end{equation}

\noindent \textbf{Scene-dependent $\mathbf{w}$ Predictor.} The predictor $\psi$ maps the scene complexity descriptor $\boldsymbol\rho$ and a
low-resolution pixel summary to a scene-dependent color axis through a CNN+MLP.
The image is first reduced by adaptive average pooling (AAP) to a fixed
$32\!\times\!32$ tensor and encoded by a 2-layer strided CNN:

\begin{equation}
\mathbf{f}_{\text{CNN}} = \mathrm{Flatten \,} \!\Bigl(\mathrm{GAP}\bigl(\mathrm{CNN}_2\!\circ\!\mathrm{ \, CNN}_1\!\circ\!\mathrm{AAP}_{32\times32}(\mathbf{I}) \, \bigr)\!\Bigr)\;\in\mathbb{R}^{32},
\end{equation}

\noindent with each $\mathrm{CNN}_k = \mathrm{Conv}_{3\times3,\,\text{stride}\,2}\!\to\!\mathrm{BN}\!\to\!\mathrm{GELU}$ at channel widths $3\!\to\!16\!\to\!32$ and a final global average pool $(\mathrm{GAP})$. The 32D CNN embedding is concatenated with the 8D scene complexity features $\boldsymbol\rho$ and passed through a 3-layer MLP of hidden width 64 to produce a logit residual $\mathbf{r}\in\mathbb{R}^{3}$. The color axis is

\begin{equation}
\mathbf{w}(\mathbf{I}) = \mathrm{softmax\,}\!\bigl(\frac{1}{\tau}(\boldsymbol\beta + \mathbf{r})\bigr) \;\in\Delta^{2},
\end{equation}

\noindent where $\boldsymbol\beta\in\mathbb{R}^{3}$ is a learnable dataset-level prior
(initialized to zero to yield uniform $\mathbf{w}=(1/3,1/3,1/3)$) and $\tau$ controls the softmax
peakedness. With only 12k parameters, $\psi$ is deliberately
lightweight: its capacity is restricted to the coarse "which channel to
suppress" mapping that LCS theoretically requires.



\subsection{SAFE Full Pipeline}

Our proposed model for pure color constancy is trained in four phases: an initial backbone pre-training, two intermediate phases that warm-start the LCS predictor, and a final joint fine-tune.

\vspace{1pt}

\noindent \textbf{Phase 1: SAFE backbone (teacher training).} Phase 1 trains the modulation network $\phi$ jointly with a
coarse residual $w$-module under the angular error
loss $\mathcal{L}_{\text{ang}}$. The feature pipelines are deterministic and serve as fixed inputs to $\phi$: the color-blind
subset $\mathbf{g}$ of the scene descriptor drives the scene-conditioned attention $\boldsymbol\alpha(\mathbf{g})$,
and the four-token illumination features
$\mathbf{f_{illum}}$ feed the residual MLP. We refer to the resulting trained
network as the \textbf{S}cene-\textbf{A}ware \textbf{F}eatur\textbf{E} Modulation network for pure color constancy (SAFE) backbone,
denoted $\phi^{(1)}$.


During Phase 1 the LCS color axis is realized by a residual $w$-module that mainly learns a global, dataset-level
axis. This lets the backbone learn the scene-conditioned attention and the residual MLP without having to simultaneously resolve a scene-dependent color axis. The frozen $\phi^{(1)}$ then serves as the teacher model in Phases 2
and 3.

\vspace{1pt}

\noindent \textbf{Phase 2: Scene-dependent w Search.} With $\phi^{(1)}$ frozen, for each training image $\mathbf{I}_i$ paired with its
ground-truth illuminant $\mathbf{y}_i^\ast$, we solve for the scene-dependent color axis that minimizes the angular error loss of $\phi^{(1)}$'s prediction. This  optimization is performed on the unconstrained logits
$\boldsymbol\ell\in\mathbb{R}^{3}$ using Adam~\cite{kingma2015adam} for 80 steps. The resulting $\mathbf{w}$ collection characterizes the scene-dependent performance ceiling that LCS can in principle achieve given
$\phi^{(1)}$.

\vspace{1pt}

\noindent \textbf{Phase 3: LCS Predictor (Student training).} With the $\mathbf{w}$ collection in hand, we pretrain the LCS predictor, denoted $\psi^{(3)}$, to imitate them. The trained $\psi^{(3)}$ generates a scene-dependent color axis for each image to re-parameterize the chromaticity normalization for subsequent forward passes, and thus determines the four-token illumination feature that the SAFE backbone consumes.

\vspace{1pt}

\noindent \textbf{Phase 4: Joint training.} Phase 4 unfreezes $\phi^{(1)}$ and $\psi^{(3)}$ and fine-tunes them
end-to-end. At every forward pass,
the scene-dependent axis $\mathbf{w} \!=\! \psi(\boldsymbol\rho, \mathbf{I})$
is used to recompute the four-token illumination representation under the $w$-weighted chromaticity, so the angular-loss gradient flows back through both the SAFE backbone {and} the LCS predictor. The objective combines the angular error loss with a predictive-consistency regularizer that discourages $\psi$ from
producing volatile $\mathbf{w}$ under small pixel perturbations:

\vspace{-0.1cm}

\begin{equation}
\mathcal{L}_{\text{cons}} \;=\; \bigl\|\,\psi(\boldsymbol\rho,\mathbf{I})\;-\;\psi\bigl(\boldsymbol\rho,\,\mathbf{I}\odot(1+\boldsymbol\eta)\bigr)\,\bigr\|_{2}^{2},\ \
\boldsymbol\eta_{p,c} \stackrel{\text{i.i.d.}}{\sim} \mathcal{N}(0,\sigma_\eta^{2}), 
\end{equation}

\noindent where the noise $\boldsymbol\eta$ is sampled per pixel and channel, applied multiplicatively so that each pixel's intensity is jittered by a factor $1+\boldsymbol\eta_{p,c}$ around its original value. The scene
descriptor $\boldsymbol\rho$ is held fixed at its value on the
unperturbed image. The full Phase 4 loss function is formulated as in Eq. (17):

\vspace{-0.1cm}

\begin{equation}
\mathcal{L}_{\text{p4}} \;=\; \mathcal{L}_{\text{ang}}\bigl(\hat{\mathbf{y}},\mathbf{y}^\ast\bigr) \;+\; \lambda_c\,\mathcal{L}_{\text{cons}}.
\end{equation}

%% file: 05_experiment.tex
\section{Experiments}
\label{sec:experiments}

\subsection{Experimental Setup}
\label{ssec:exp_setup}

We evaluate SAFE on the PolyU Pure Color Dataset V2~\cite{Liu2026Color}. It contains 1,271 pure-color and near-monochromatic images, providing a rigorous benchmark for color constancy in color-deficient scenarios. All images are resized to a size of $512\times512$. We use the standard angular error $\theta$ (expressed in degrees) to measure the deviation between the estimated illuminant vector and the ground-truth vector. A lower angular error signifies superior estimation accuracy. Following established protocols, we adopt a three-fold cross-validation scheme. In each fold, two-thirds of the data are used for training while the remainder serves for evaluation. We report a comprehensive results from the angular error distribution, including the mean, median, trimean, and the means of the best 25\% and worst 25\% subsets. All learning-based methods were retrained from scratch using officially released code under same experimental conditions.

\input{PolyUv2Results}
\input{GehlerResults}

\subsection{Implementation Details}
\label{ssec:exp_implement}

Our proposed model is trained in the four phases described in Sec.~\ref{sec:method}; per-phase hyperparameters are
listed below. All neural network optimization uses AdamW~\cite{loshchilov2019adamw} with $\beta_1=0.9$,
$\beta_2=0.999$, a maximum of 500 epochs, and early stopping on
validation mean angular error (patience 40). All training and evaluation runs were conducted on a single NVIDIA RTX 5090 GPU.

\vspace{1pt}

\noindent \textbf{Architecture.} The modulation network $\phi$ has hidden
width $d{=}80$, $L{=}7$ residual blocks, and MLP hidden width
$d_g{=}4$ ($49{,}343$ parameters). The LCS predictor $\psi$ consists
of a 2-layer strided CNN (channel widths $3{\to}16{\to}32$) followed
by a 3-layer MLP of hidden width 64 ($12{,}166$ parameters). The full
pipeline has $\sim\!61.5\,$K trainable parameters.

\vspace{1pt}

\noindent \textbf{Phase 1: SAFE backbone.} Hyperparameters: learning rate $1.47{\times}10^{-3}$,
weight decay $2.54{\times}10^{-4}$, gradient-clip norm $0.42$,
dropout $0.091$, color-space LR multiplier $0.205$, $w$-module
temperature $\tau{=}1.21$, linear LR schedule decaying to $34.7\%$ of
the initial rate, batch size $64$.

\vspace{1pt}

\noindent \textbf{Phase 2: Scene-dependent w search.} With $\phi^{(1)}$
frozen, we solve per-image for the LCS logits that minimise the
teacher's angular error using $80$ steps of Adam with learning rate $0.05$, processed in batches of $64$ images. The search is offline and runs once per fold.

\vspace{1pt}

\noindent \textbf{Phase 3: LCS Predictor (Student training).} Hyperparameters: up to $200$ epochs (patience $80$), batch size $64$,
learning rate $1{\times}10^{-3}$, weight decay $1{\times}10^{-4}$,
cosine LR schedule decaying to $10\%$ of the initial rate.

\vspace{1pt}

\noindent \textbf{Phase 4: Joint training.} Hyperparameters: learning rate $1.93{\times}10^{-4}$, weight
decay $4.01{\times}10^{-4}$, gradient-clip norm $0.22$, dropout
$0.125$, color-space LR multiplier $0.368$, $\tau{=}2.24$, cosine LR
schedule decaying to $27.1\%$ of the initial rate, consistency weight
$\lambda_c{=}0.164$, consistency noise $\sigma_\eta{=}0.086$, batch
size $32$.


\input{wAnalysis}
\input{Comparison}

\subsection{Results and Comparisons}
\label{ssec:results}

As in Table~\ref{tab:polyu}, SAFE (P1) denotes the SAFE backbone trained with the coarse residual w-module (a near-global LCS) in the phase 1; SAFE (P4) denotes the full pipeline after joint fine-tuning of the SAFE backbone with the scene-dependent LCS predictor in the phase 4. Our method achieves superior performance on the PolyU Pure Color dataset V2, outperforming state-of-the-art methods in the mean, median, tri-mean, best 25\%, and worst 25\% angular error metrics. Fig.~\ref{fig:comparison} shows qualitative results on several pure-color scenes from the PolyU V2 dataset. Furthermore, the integration of scene-dependent LCS enhances the accuracy of illuminant estimation, reducing the mean error from 2.12° to 1.94°.

The LCS is theoretically
motivated by chromaticity collapse on pure-color scenes.
We verify empirically that the learned predictor $\psi$ has
internalized this prior. After Phase 1, the coarse residual $w$-module converges to a very mild,
dataset-level perturbation of the uniform axis. Averaged across the
three PolyU V2 folds, the global axis is $\mathbf{w}_{\mathrm{glob}} \!\approx\! (0.325,\,0.319,\,0.356)$,
with a per-component standard deviation of at most $0.004$ across folds. For each test image, we compute the scene-dependent axis deviation
$\Delta_w \!=\, \rVert \,\mathbf{w}_{\text{pred}} -
\mathbf{w}_{\text{global}} \, \rVert_2$
(deviation from the fold-level global axis) and plot it against scene
entropy $H$. Both the LOWESS (Locally Weighted Scatterplot Smoothing) curve with a $95\%$ bootstrap confidence band and the hexbin density overlaid
with LOWESS shown in Fig.~\ref{fig:wAnalysis} reveal a consistent decreasing trend: pure-color scenes drive significantly larger axis deviations than diverse ones. These results show that the LCS predictor $\psi$ allocates larger scene-dependent axis corrections to pure-color scenes, which is exactly the regime the LCS is designed to address.


\input{inference}


To also evaluate SAFE in general scenes, we report its performance on the Gehler-Shi dataset~\cite{gehler2008bayesian,shi2010reprocessing} in Table~\ref{tab:gehler_results}. The results demonstrate that despite its specialization in pure-color scenes, SAFE and the LCS framework still achieve comparable performance in general scenarios.

\subsection{Runtime Analysis}
\label{ssec:exp_cost}

We measure parameter count and
inference latency (ms) for several learning-based baselines shown in Table~\ref{tab:inference}. Among them, our model achieves the highest accuracy of illuminant estimation while requiring fewer parameters and
fewer latency than most approaches, with only a modest increase in both
quantities over the two smallest baseline (CCC and ePCC).

\subsection{Ablation Study}
\label{ssec:ablation}

\input{FeatureImportance}
\input{Ablation_p1}

Table~\ref{tab:ablation_p1} reports the five angular error metrics for the scene-routing modulation and four token ablations for SAFE backbone (Phase 1), validating that all components are essential for superior illuminant estimation in pure-color scenes.

We also verify whether the network's learned preferences are consistent with the ablation ranking. Fig.~\ref{fig:fImportance} illustrates three signals calculated from the trained SAFE backbone (phase 1): the modulation attention mask $\bar{\alpha}_j$, the
input projection weight $\lVert W_0[:,j]\rVert_2$, and their
product $\bar{\alpha}_j \!\cdot\! \lVert W_0[:,j]\rVert_2$ as the
effective contribution. Each signal is aggregated by per-token mean to neutralize the differing token sizes. All three signals produce the same ranking, ${C}\!>\!{B}\!>\!{A}\!>\!{D}$, aligned with the empirical ablation ordering demonstrated in Table~\ref{tab:ablation_p1}: ($\Delta\overline{\text{MAE}}$: C ${+}0.22^{\circ}$,
B ${+}0.10^{\circ}$, A ${+}0.08^{\circ}$, D ${+}0.03^{\circ}$).


%% file: PolyUv2Results.tex
\definecolor{lightred}{RGB}{253,191,191}
\definecolor{lightorange}{RGB}{255,223,191}
\definecolor{lightyellow}{RGB}{254,240,198}

\begin{table}[t]
\setlength{\abovecaptionskip}{3pt}
\centering
\small
\caption{\textbf{Evaluation results on the PolyUv2 pure color dataset.} We report the mean angular error (in degrees) based on 3-fold cross-validation. SAFE outperforms existing methods, yielding the lowest mean, median, tri-mean, best-25\%, and worst-25\% angular errors, underscoring its superior stability.}

\label{tab:polyu}
\begin{tabular}{l|ccccc}

\toprule
\makebox[3cm][l]{Method} &
\makebox[1.1cm]{Mean} &
\makebox[1.1cm]{Med.} &
\makebox[1.1cm]{Tri.} &
\makebox[1.1cm]{B-25\%} &
\makebox[1.1cm]{W-25\%} \\
\midrule

WP ~\cite{land1977retinex}
& 5.99
& 4.37 
& 4.72
& 0.89 
& 14.08 \\

GW ~\cite{buchsbaum1980spatial}
& 6.78
& 5.53 
& 5.69
& 1.73 
& 14.08 \\

GE ~\cite{van2007edge}
& 4.73
& 3.38 
& 3.80
& 1.11
& 10.52 \\

SoG ~\cite{finlayson2004shades}
& 5.55
& 4.50 
& 4.74
& 1.05 
& 11.84 \\

LSRS ~\cite{Gao2014EfficientCC}
& 6.00
& 5.02
& 5.10
& 1.46
& 12.52 \\

GP ~\cite{Yang2015Efficient}
& 4.56 
& 3.51 
& 3.84
& 0.82 
& 10.01 \\

GI ~\cite{qian2019finding}
& 4.09
& 2.81
& 3.17
& 0.67
& 9.65 \\

FC$^4$ ~\cite{hu2017fc4}
& 2.48
& 1.79
& 2.05
& 0.55
& 5.57

\\

$C^{4}$ ~\cite{yu2020cascading}
& 2.23
& 1.49
& 1.75
& 0.51
& 5.05 \\

$C^5$ ~\cite{afifi2021cross}
& 2.18
& 1.55
& 1.78
& 0.52
& \cellcolor{lightyellow}4.95 \\

PCC ~\cite{Wei2023ColorCF}
& 2.39
& 1.60
& 1.76
& 0.51
& 5.47 \\

ePCC ~\cite{Liu2026Color}
& 2.16 
& \cellcolor{lightyellow}1.39 
& 1.61
& 0.45 
& 5.08 \\

\midrule
SAFE w/o LCS
& \cellcolor{lightyellow}2.12
& 1.44
& \cellcolor{lightyellow}1.56
& \cellcolor{lightyellow}0.43
& 5.01 \\

SAFE w/ LCS (P1)
& \cellcolor{lightorange}2.03
& \cellcolor{lightorange}1.38
& \cellcolor{lightorange}1.50
& \cellcolor{lightorange}0.39
& \cellcolor{lightorange}4.80 \\

SAFE w/ LCS (P4)
& \cellcolor{lightred}1.94
& \cellcolor{lightred}1.27
& \cellcolor{lightred}1.39
& \cellcolor{lightred}0.36
& \cellcolor{lightred}4.66 \\
\bottomrule
\end{tabular}
\end{table}

%% file: GehlerResults.tex
\definecolor{lightred}{RGB}{253,191,191}
\definecolor{lightorange}{RGB}{255,223,191}
\definecolor{lightyellow}{RGB}{254,240,198}

\begin{table}[t]
\setlength{\abovecaptionskip}{3pt}
\centering
\small
\caption{\textbf{Evaluation results on the Gehler-Shi dataset.} We report the mean angular error (in degrees) based on 3-fold cross-validation. The results prove that SAFE is not limited to pure-color scenes, but achieves comparable performance on general scenes.}

\label{tab:gehler_results}
\begin{tabular}{l|ccccc}
\toprule
\makebox[3cm][l]{Method} &
\makebox[1.1cm]{Mean} &
\makebox[1.1cm]{Med.} &
\makebox[1.1cm]{Tri.} &
\makebox[1.1cm]{B-25\%} &
\makebox[1.1cm]{W-25\%} \\
\midrule

WP ~\cite{land1977retinex}
& 7.55 
& 5.68 
& 6.35
& 1.45 
& 16.12 \\

GW ~\cite{buchsbaum1980spatial}
& 6.36 
& 6.28 
& 6.28
& 2.33 
& 10.58 \\

SoG ~\cite{finlayson2004shades} 
& 4.93 & 4.01 & 4.23 & 1.14 & 10.20\\

PCA-CC ~\cite{cheng2014illuminant}
& 3.52 & 2.14 & 2.47 & 0.50 & 8.74 \\

Woo \textit{et al.} ~\cite{woo2017improving}
& 4.30 & 2.86 & 3.31 & 0.71 & 10.14 \\

GI ~\cite{qian2019finding}
& 3.07 & 1.87 & 2.16 & 0.43 & 7.62 \\

$C^{4}$ ~\cite{yu2020cascading}
& \cellcolor{lightred}1.35
& \cellcolor{lightred}0.88
& \cellcolor{lightred}0.99
& \cellcolor{lightorange}0.28
& \cellcolor{lightred}3.21 \\

CLCC ~\cite{lo2021clcc}
& \cellcolor{lightorange}1.44
& \cellcolor{lightorange}0.92
& \cellcolor{lightorange}1.04
& \cellcolor{lightred}0.27
& \cellcolor{lightyellow}3.48 \\

GCC ~\cite{chang2025gcc}
& \cellcolor{lightyellow}1.91 
& 1.80 
& 1.84
& 0.60 
& \cellcolor{lightorange}3.46 \\

PCC ~\cite{Wei2023ColorCF}
& 2.64 
& 1.61 
& 1.83
& 0.45 
& 6.62 \\

ePCC ~\cite{Liu2026Color}
& 2.45 
& 1.59 
& 1.78
& 0.44 
& 5.87 \\

\midrule

SAFE w/o LCS 
& 2.35
& 1.69
& 1.78
& 0.47
& 5.57 \\

SAFE w/LCS (P1)
& 2.33
& 1.66
& 1.75
& 0.47
& 5.43 \\

SAFE w/ LCS (P4)
& 2.23
& \cellcolor{lightyellow}1.56
& \cellcolor{lightyellow}1.69
& \cellcolor{lightyellow}0.41
& 5.24 \\
\bottomrule
\end{tabular}
\end{table}

%% file: wAnalysis.tex
\begin{figure*}[tp]
    \centering
    \includegraphics[width=\linewidth, trim={0cm 0cm 0cm 0cm},clip]{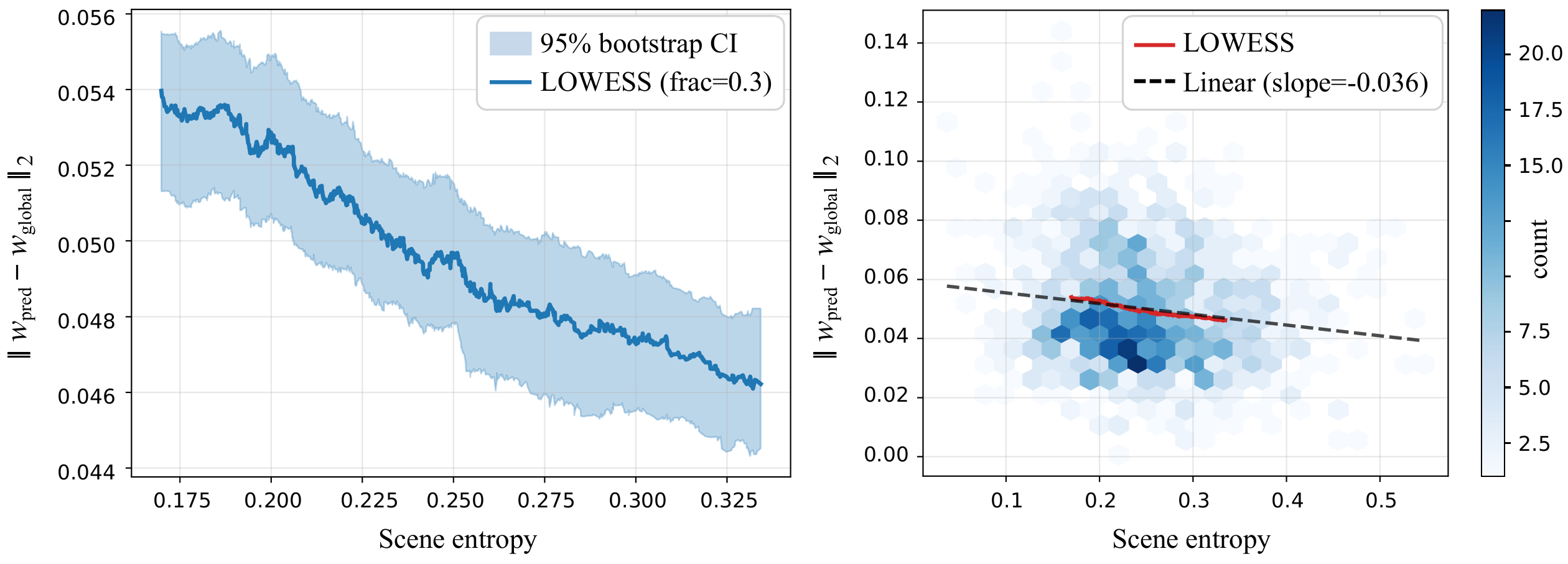}
\vspace{-5mm}
    \caption{\textbf{Scene-dependent LCS deviation versus scene entropy.} {(Left figure)} The LOWESS smoother with $95\%$ bootstrap confidence band. {(Right figure)} Hexbin density of the same data with the LOWESS smoother overlaid in red. Quantitatively, the lowest-entropy quintile ($n{=}255$) attains $\bar{\Delta}_w = 0.0547$ versus $0.0462$ for the highest-entropy
    quintile ($18\%$ relative increase on pure-color scenes) and the Pearson correlation $r(H,\Delta_w) = -0.149$ is highly significant ($p < 10^{-7}$, $N{=}1271$). The analysis validates that the LCS design further improves the accuracy of illumination estimation in pure-color scenes.}
    \label{fig:wAnalysis}
\end{figure*}

%% file: Comparison.tex
\begin{figure*}[!t]
    \centering
    \includegraphics[width=\linewidth, trim={0cm -0.1cm 0cm 0cm},clip]{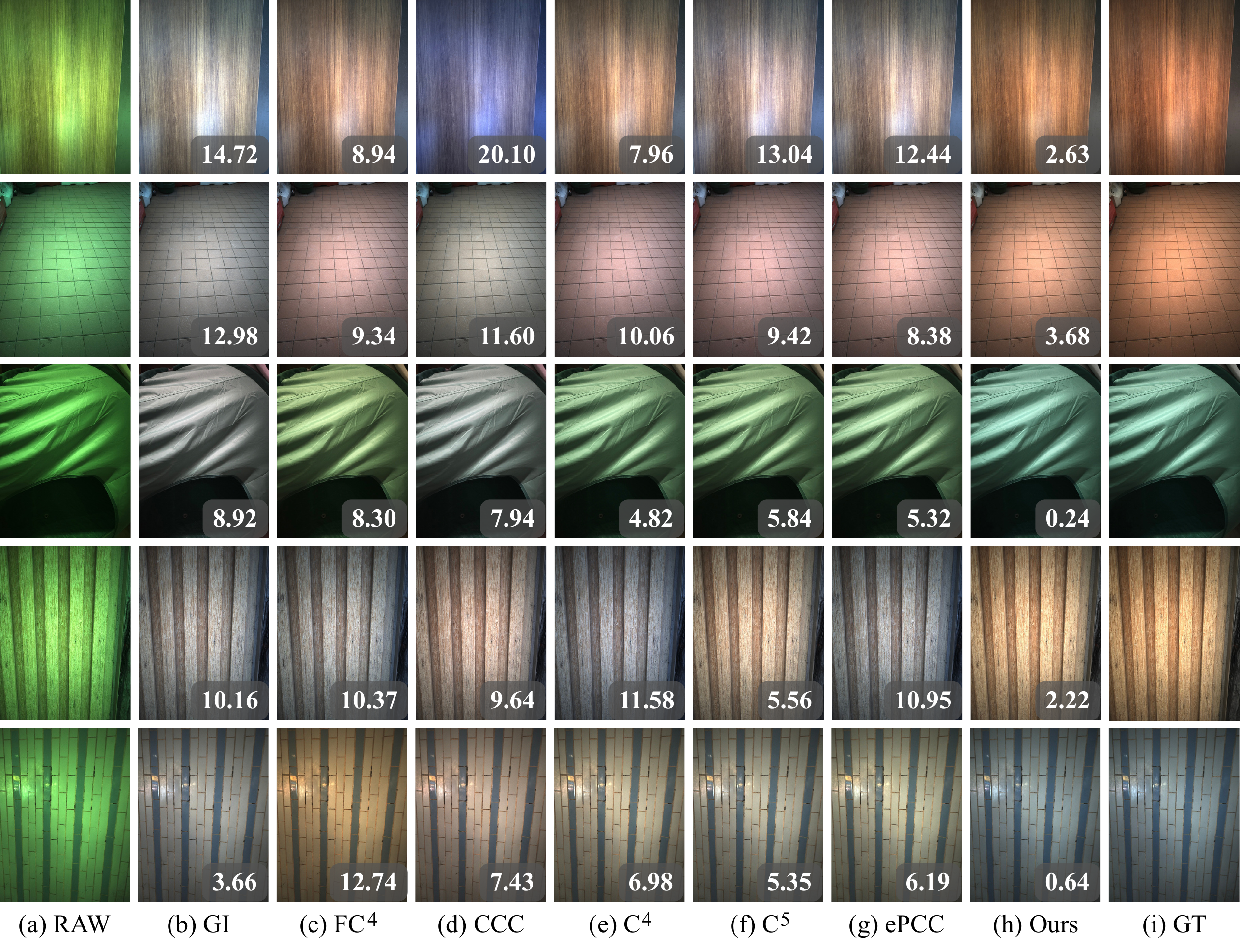}
\vspace{-5mm}
    \caption{\textbf{Qualitative  comparison on images from the PolyU Pure Color dataset V2.} Images are gamma-corrected for visualization. The angular error (in degree) is reported in the bottom-right corner of each image.}
    \label{fig:comparison}
\end{figure*}

%% file: inference.tex
\begin{table}[t]
\setlength{\abovecaptionskip}{3pt}
\centering
\small
\caption{\textbf{Inference cost on PolyU V2 dataset.} We report the parameters and median/min latency (ms) for several learning-based baselines. All measurements use batch=1, matching the standard single-image inference setting of the benchmark. Our model achieves the best illumination estimation accuracy. It is smaller in both parameters and latency than every baseline except CCC and ePCC, over which it incurs only a
moderate increase.}

\label{tab:inference}
\begin{tabular}{w{l}{3.15cm}|w{c}{1.4cm}w{c}{1.4cm}w{c}{1.4cm}w{c}{1.4cm}w{c}{1.4cm}w{c}{1.4cm}}

\toprule
{Inference cost} &
{CCC~\cite{barron2015convolutional}} &
{FC$^4$~\cite{hu2017fc4}} &
{$C^{4}$~\cite{yu2020cascading}} &
{$C^5$~\cite{afifi2021cross}} &
{ePCC~\cite{Liu2026Color}} &
{SAFE} 
\\
\midrule

Parameters 
& 700 
& 1.71M
& 5.12M 
& 412k
& 514
& 61.5k \\

Median/Min Latency
& 0.24/0.23
& 0.95/0.89 
& 2.58/2.30
& 5.73/4.53 
& 0.17/0.16
& 0.66/0.61 \\

\bottomrule
\end{tabular}
\end{table}

%% file: FeatureImportance.tex
\begin{figure*}[tp]
    \centering
    \includegraphics[width=\linewidth, trim={0cm 0cm 0cm 0cm},clip]{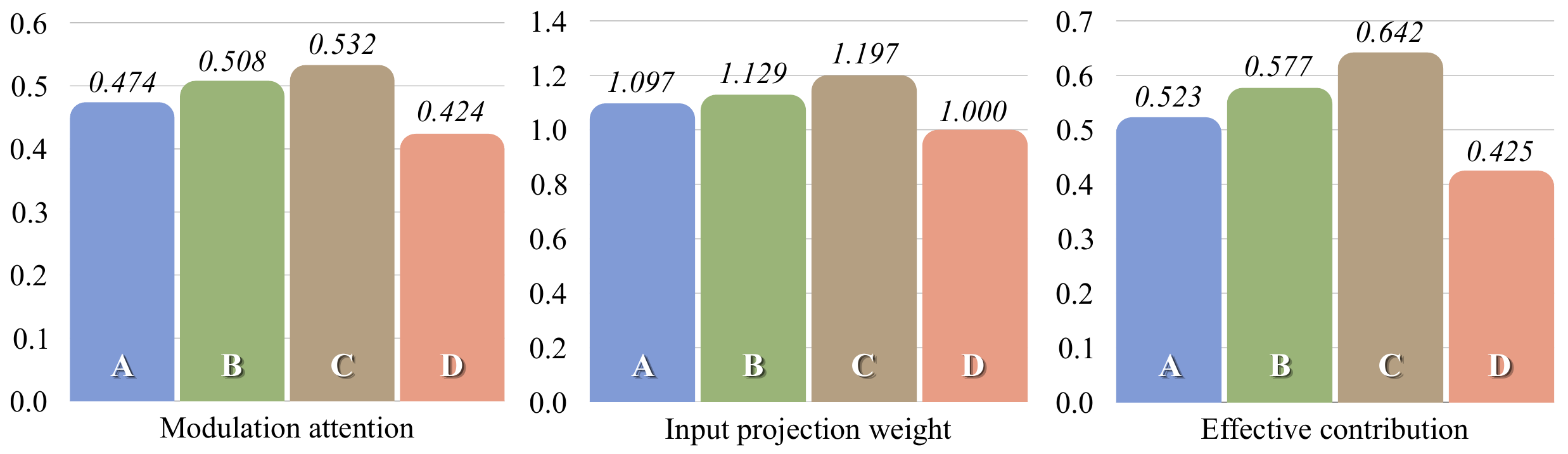}
\vspace{-5mm}
    \caption{\textbf{Introspection on the SAFE backbone.} The modulation attention $\alpha(\mathbf{g})$ quantifies how strongly the modulation network preserves each feature dimension; the input projection weight $\lVert W_0[:,j]\rVert_2$ quantifies how much linear weight the backbone allocates to each dimension; and the intrinsic effective contribution quantifies how important each token is to the illumination estimation.}
    \label{fig:fImportance}
\end{figure*}

%% file: Ablation_p1.tex
\begin{table*}[t]
\setlength{\abovecaptionskip}{3pt}
\centering
\small
\caption{Ablation studies on different impact of the tokens and the scene-routing modulation module (SRM). The best performance is reached when all the four tokens and the modulation are employed, validating the design of our framework.} 
\label{tab:ablation_p1}

\begin{tabular}{ccccc|ccccc}
\toprule
\makebox[1.13cm]{ \text{SRM}} &
\makebox[1.13cm]{ \textbf{A}\text{(8D)}} &
\makebox[1.13cm]{ \textbf{B}\text{(8D)}} & 
\makebox[1.13cm]{ \textbf{C}\text{(6D)}} & 
\makebox[1.13cm]{ \textbf{D}\text{(2D)}} & 
\makebox[1.10cm]{Mean} & \makebox[1.10cm]{Med.} & \makebox[1.10cm]{Tri.} & \makebox[1.10cm]{B-25\%} & \makebox[1.10cm]{W-25\%} \\
\midrule
- & 
\checkmark & 
\checkmark & 
\checkmark & 
\checkmark &
2.08 & 
1.38 & 
1.53 & 
0.43 & 
4.96
\\
\checkmark &
- &  
\checkmark & 
\checkmark & 
\checkmark &
2.11 & 
1.43 & 
1.56 & 
0.43 & 
4.97
\\
\checkmark &
\checkmark & 
- & 
\checkmark &
\checkmark & 
2.13 & 
1.42 & 
1.56 & 
0.40 & 
5.09 
\\
\checkmark &
\checkmark & 
\checkmark & 
- &
\checkmark &
2.25 & 
1.52 & 
1.68 & 
0.47 &
5.30
\\
\checkmark &
\checkmark & 
\checkmark & 
\checkmark & 
- &
2.06 & 
1.43 & 
1.53 & 
0.43 &
4.89 
\\
\checkmark &
\checkmark & 
\checkmark & 
\checkmark &
\checkmark & 
\cellcolor{black!10}\textbf{2.03} & 
\cellcolor{black!10}\textbf{1.38} & 
\cellcolor{black!10}\textbf{1.50} & 
\cellcolor{black!10}\textbf{0.39} & 
\cellcolor{black!10}\textbf{4.80} 
\\

\bottomrule
\end{tabular}
\end{table*}

%% file: 10_conclusion.tex
\section{Conclusion}
\label{sec:conclusion}

We propose a color constancy framework for pure-color scenes through three coordinated components: an interpretable \emph{four-token} semantic decomposition of illumination features, a scene-conditioned modulation network (\emph{SAFE}) that selectively reweights these tokens, and a \emph{Learned Color Space} (LCS), a scene-dependent chromaticity
normalization that restores feature discriminability. Combining these novel designs, our model achieves the highest estimation accuracy compared to state-of-the-art baselines in pure-color scenes. As a feature-driven, lightweight model, SAFE offers a computationally efficient solution for practical, real-time deployment in mobile cameras, autonomous vehicles, and other embedded systems where computational resources are limited.

\vspace{1pt}

\noindent \textbf{Limitation and Future Work.} The proposed LCS framework is naturally sensor-sensitive: chromaticity normalization reweighting can entangle with sensor spectral sensitivities. Future work will involve exploring a sensor-aware LCS that conditions the predictor on a sensor characteristic descriptor, allowing a single predictor to deploy across multiple cameras without per-sensor retraining.


%% file: main.bbl
\begin{thebibliography}{10}
\providecommand{\url}[1]{\texttt{#1}}
\providecommand{\urlprefix}{URL }
\providecommand{\doi}[1]{https://doi.org/#1}

\bibitem{afifi2021cross}
Afifi, M., Barron, J.T., LeGendre, C., Tsai, Y.T., Bleibel, F.: Cross-camera
  convolutional color constancy. In: IEEE/CVF Int. Conf. Computer Vision. pp.
  1981--1990 (2021)

\bibitem{afifi2020deep}
Afifi, M., Brown, M.S.: Deep white-balance editing. In: IEEE/CVF Conf. Computer
  Vision and Pattern Recognition. pp. 1397--1406 (2020)

\bibitem{afifi2022auto}
Afifi, M., Brubaker, M.A., Brown, M.S.: Auto white-balance correction for
  mixed-illuminant scenes. In: IEEE/CVF Winter Conf. Applications of Computer
  Vision. pp. 1210--1219 (2022)

\bibitem{afifi2019color}
Afifi, M., Price, B., Cohen, S., Brown, M.S.: When color constancy goes wrong:
  Correcting improperly white-balanced images. In: Proceedings of the IEEE/CVF
  conference on computer vision and pattern recognition. pp. 1535--1544 (2019)

\bibitem{barron2015convolutional}
Barron, J.T.: Convolutional color constancy. In: IEEE Int. Conf. Computer
  Vision. pp. 379--387 (2015)

\bibitem{barron2017fast}
Barron, J.T., Tsai, Y.T.: Fast fourier color constancy. In: IEEE/CVF Conf.
  Computer Vision and Pattern Recognition. pp. 886--894 (2017)

\bibitem{buchsbaum1980spatial}
Buchsbaum, G.: A spatial processor model for object colour perception. Journal
  of the Franklin Institute  \textbf{310}(1),  1--26 (1980)

\bibitem{chang2025gcc}
Chang, C.W., Fan, C.D., Chang, C.C., Lo, Y.C., Tseng, Y.C., Huang, J.L., Liu,
  Y.L.: Gcc: Generative color constancy via diffusing a color checker. In:
  IEEE/CVF Conf. Computer Vision and Pattern Recognition. pp. 10868--10878
  (2025)

\bibitem{cheng2024nighttime}
Cheng, C., Yang, K.F., Wan, X.M., Chan, L.L.H., Li, Y.J.: Nighttime color
  constancy using robust gray pixels. J. Opt. Soc. Am. A  \textbf{41}(3),
  476--488 (2024)

\bibitem{cheng2014illuminant}
Cheng, D., Prasad, D.K., Brown, M.S.: Illuminant estimation for color
  constancy: why spatial-domain methods work and the role of the color
  distribution. J. Opt. Soc. Am. A  \textbf{31}(5),  1049--1058 (2014)

\bibitem{cheng2026perception}
Cheng, Y., Cui, Z., Gu, L., Su, S., Zhang, Z.: Perception-inspired color space
  design for photo white balance editing. In: Proceedings of the IEEE/CVF
  Winter Conference on Applications of Computer Vision. pp. 3741--3749 (2026)

\bibitem{ebner2004parallel}
Ebner, M.: A parallel algorithm for color constancy. Journal of Parallel and
  Distributed Computing  \textbf{64}(1),  79--88 (2004)

\bibitem{entok2024pixel}
Entok, U.C., Laakom, F., Pakdaman, F., Gabbouj, M.: Pixel-wise color constancy
  via smoothness techniques in multi-illuminant scenes. In: 2024 IEEE
  International Conference on Image Processing (ICIP). pp. 2737--2743. IEEE
  (2024)

\bibitem{farghaly2023two}
Farghaly, M., Mansour, R.F., Sewisy, A.A.: Two-stage deep learning framework
  for srgb image white balance. Signal, Image and Video Processing
  \textbf{17}(1),  277--284 (2023)

\bibitem{finlayson2004shades}
Finlayson, G.D., Trezzi, E.: Shades of gray and colour constancy. In: Color and
  Imaging Conference. vol.~12, pp. 37--41. Society of Imaging Science and
  Technology (2004)

\bibitem{Gao2014EfficientCC}
Gao, S., Han, W., Yang, K., Li, C., Li, Y.: Efficient color constancy with
  local surface reflectance statistics. In: European Conf. Computer Vision. pp.
  159--173 (2014)

\bibitem{gehler2008bayesian}
Gehler, P.V., Rother, C., Blake, A., Minka, T., Sharp, T.: Bayesian color
  constancy revisited. In: IEEE/CVF Conf. Computer Vision and Pattern
  Recognition. pp.~1--8. IEEE (2008)

\bibitem{gijsenij2010generalized}
Gijsenij, A., Gevers, T., Van De~Weijer, J.: Generalized gamut mapping using
  image derivative structures for color constancy. Int. J. Computer Vision
  \textbf{86},  127--139 (2010)

\bibitem{hashemi2010image}
Hashemi, S., Kiani, S., Noroozi, N., Moghaddam, M.E.: An image contrast
  enhancement method based on genetic algorithm. Pattern Recognition Letters
  \textbf{31}(13),  1816--1824 (2010)

\bibitem{hendrycks2016gaussian}
Hendrycks, D., Gimpel, K.: Gaussian error linear units (gelus). arXiv preprint
  arXiv:1606.08415  (2016)

\bibitem{hu2017fc4}
Hu, Y., Wang, B., Lin, S.: Fc4: Fully convolutional color constancy with
  confidence-weighted pooling. In: IEEE/CVF Conf. Computer Vision and Pattern
  Recognition. pp. 4085--4094 (2017)

\bibitem{jiang2012flexible}
Jiang, T., Nguyen, D., Kuhnert, K.D.: A flexible auto white balance based on
  histogram overlap. In: Asian Conference on Computer Vision. pp. 170--181.
  Springer (2012)

\bibitem{jiang2012auto}
Jiang, T., Nguyen, D., Kuhnert, K.D.: Auto white balance using the coincidence
  of chromaticity histograms. In: 2012 Eighth International Conference on
  Signal Image Technology and Internet Based Systems. pp. 201--208. IEEE (2012)

\bibitem{jiao2026efficient}
Jiao, W., Li, B., Shang, W., Wang, P., Ren, D.: Efficient raw image deblurring
  with adaptive frequency modulation. Advances in Neural Information Processing
  Systems  \textbf{38},  70191--70223 (2026)

\bibitem{kim2025ccmnet}
Kim, D., Afifi, M., Kim, D., Brown, M.S., Kim, S.J.: Ccmnet: Leveraging
  calibrated color correction matrices for cross-camera color constancy. arXiv
  preprint arXiv:2504.07959  (2025)

\bibitem{kim2021large}
Kim, D., Kim, J., Nam, S., Lee, D., Lee, Y., Kang, N., Lee, H.E., Yoo, B., Han,
  J.J., Kim, S.J.: Large scale multi-illuminant (lsmi) dataset for developing
  white balance algorithm under mixed illumination. In: Proceedings of the
  IEEE/CVF International Conference on Computer Vision. pp. 2410--2419 (2021)

\bibitem{kim2024attentive}
Kim, D., Kim, J., Yu, J., Kim, S.J.: Attentive illumination decomposition model
  for multi-illuminant white balancing. In: Proceedings of the IEEE/CVF
  Conference on Computer Vision and Pattern Recognition. pp. 25512--25521
  (2024)

\bibitem{kim2023paramisp}
Kim, W., Kim, G., Lee, J., Lee, S., Baek, S.H., Cho, S.: Paramisp: Learned
  forward and inverse isps using camera parameters. arXiv preprint
  arXiv:2312.13313  (2023)

\bibitem{kingma2015adam}
Kingma, D.P., Ba, J.: Adam: A method for stochastic optimization. In:
  International Conference on Learning Representations (ICLR) (2015)

\bibitem{land1977retinex}
Land, E.H.: The retinex theory of color vision. Scientific American
  \textbf{237}(6),  108--129 (1977)

\bibitem{Lee2026RLAWB}
Lee, Y.K., Chen, K.L., Chang, C.C., Liu, Y.L.: Rl-awb: Deep reinforcement
  learning for auto white balance correction in low-light night-time scenes.
  arXiv preprint arXiv:2601.05249  (2026)

\bibitem{lee2024efficient}
Lee, Y.K., Ding, J.J.: Efficient color image denoising using dwt-based noise
  estimation and adaptive wiener filter. In: 2024 8th International Conference
  on Imaging, Signal Processing and Communications (ICISPC). pp. 47--51. IEEE
  (2024)

\bibitem{li2023swbnet}
Li, C., Kang, X., Zhang, Z., Ming, A.: Swbnet: a stable white balance network
  for srgb images. In: Proceedings of the AAAI Conference on Artificial
  Intelligence. vol.~37, pp. 1278--1286 (2023)

\bibitem{li2024nightcc}
Li, S., Tan, R.T.: Nightcc: Nighttime color constancy via adaptive channel
  masking. In: IEEE/CVF Conf. Computer Vision and Pattern Recognition. pp.
  25522--25531 (2024)

\bibitem{liang2022semantically}
Liang, D., Li, L., Wei, M., Yang, S., Zhang, L., Yang, W., Du, Y., Zhou, H.:
  Semantically contrastive learning for low-light image enhancement. In:
  Proceedings of the AAAI conference on artificial intelligence. vol.~36, pp.
  1555--1563 (2022)

\bibitem{Liu2026Color}
Liu, Y., Wei, M.: Color constancy from a pure color view: An edge-aware
  algorithm for a wider application. Color Research \& Application
  \textbf{51}(1),  e70036 (2026)

\bibitem{lo2021clcc}
Lo, Y.C., Chang, C.C., Chiu, H.C., Huang, Y.H., Chen, C.P., Chang, Y.L., Jou,
  K.: Clcc: Contrastive learning for color constancy. In: IEEE/CVF Conf.
  Computer Vision and Pattern Recognition. pp. 8053--8063 (2021)

\bibitem{loshchilov2019adamw}
Loshchilov, I., Hutter, F.: {Decoupled Weight Decay Regularization}. In:
  International Conference on Learning Representations (ICLR) (2019)

\bibitem{luo2026enhancing}
Luo, H., Li, R., Liang, J.: Enhancing multi-illuminant color constancy through
  multi-scale estimation and high-frequency preservation. The Visual Computer
  \textbf{42}(3), ~163 (2026)

\bibitem{mohammadi2026low}
Mohammadi, M., Honari, S., Tsogkas, S., Aumentado-Armstrong, T., Brown, M.S.,
  Mohomed, I., Derpanis, K.G., Levinshtein, A., Gilitschenski, I.: Why
  low-light cameras go color blind: Removing color bias in raw denoising. arXiv
  preprint arXiv:2607.11090  (2026)

\bibitem{Perez2018FiLM}
Perez, E., Strub, F., de~Vries, H., Dumoulin, V., Courville, A.: Film: Visual
  reasoning with a general conditioning layer. In: AAAI Conf. Artificial
  Intelligence. pp. 3942--3951 (2018)

\bibitem{qian2019finding}
Qian, Y., Kamarainen, J.K., Nikkanen, J., Matas, J.: On finding gray pixels.
  In: IEEE/CVF Conf. Computer Vision and Pattern Recognition. pp. 8062--8070
  (2019)

\bibitem{rim2022realistic}
Rim, J., Kim, G., Kim, J., Lee, J., Lee, S., Cho, S.: Realistic blur synthesis
  for learning image deblurring. In: European conference on computer vision.
  pp. 487--503. Springer (2022)

\bibitem{serrano2025revisiting}
Serrano-Lozano, D., Arora, A., Herranz, L., Derpanis, K.G., Brown, M.S.,
  Vazquez-Corral, J.: Revisiting image fusion for multi-illuminant
  white-balance correction. In: Proceedings of the IEEE/CVF International
  Conference on Computer Vision. pp. 8275--8284 (2025)

\bibitem{shafer1985using}
Shafer, S.A.: Using color to separate reflection components. Color Research \&
  Application  \textbf{10}(4),  210--218 (1985)

\bibitem{shi2010reprocessing}
Shi, L., Funt, B.: Re-processing of gehler's raw dataset. In: Simon Fraser
  University Color Research Group (2010)

\bibitem{song2019neural}
Song, A., Faugeras, O., Veltz, R.: A neural field model for color perception
  unifying assimilation and contrast. PLoS computational biology
  \textbf{15}(6),  e1007050 (2019)

\bibitem{Sun2023SAFM}
Sun, L., Dong, J., Tang, J., Pan, J.: Spatially-adaptive feature modulation for
  efficient image super-resolution. In: IEEE/CVF Int. Conf. Computer Vision.
  pp. 13190--13199 (2023)

\bibitem{tan2004color}
Tan, R.T., Nishino, K., Ikeuchi, K.: Color constancy through inverse-intensity
  chromaticity space. Journal of the Optical Society of America A
  \textbf{21}(3),  321--334 (2004)

\bibitem{tan2003illumination}
Tan, T., Nishino, K., Ikeuchi, K.: Illumination chromaticity estimation using
  inverse-intensity chromaticity space. In: 2003 IEEE Computer Society
  Conference on Computer Vision and Pattern Recognition, 2003. Proceedings.
  vol.~1, pp.~I--I. IEEE (2003)

\bibitem{tang2022transfer}
Tang, Y., Kang, X., Li, C., Lin, Z., Ming, A.: Transfer learning for color
  constancy via statistic perspective. In: AAAI Conf. Artificial Intelligence.
  vol.~36, pp. 2361--2369 (2022)

\bibitem{tian2023multi}
Tian, C., Zheng, M., Zuo, W., Zhang, B., Zhang, Y., Zhang, D.: Multi-stage
  image denoising with the wavelet transform. Pattern Recognition
  \textbf{134},  109050 (2023)

\bibitem{ulucan2023block}
Ulucan, O., Ulucan, D., Ebner, M.: Block-based color constancy: The deviation
  of salient pixels. In: ICASSP 2023-2023 IEEE International Conference on
  Acoustics, Speech and Signal Processing (ICASSP). pp.~1--5. IEEE (2023)

\bibitem{ulucan2024computational}
Ulucan, O., Ulucan, D., Ebner, M.: A computational model for color assimilation
  illusions and color constancy. In: Proceedings of the Asian Conference on
  Computer Vision. pp. 630--647 (2024)

\bibitem{van2007edge}
Van De~Weijer, J., Gevers, T., Gijsenij, A.: Edge-based color constancy. IEEE
  Trans. Image Processing  \textbf{16}(9),  2207--2214 (2007)

\bibitem{Wang2018DeepSFT}
Wang, X., Yu, K., Dong, C., Loy, C.C.: Recovering realistic texture in image
  super-resolution by deep spatial feature transform. In: IEEE/CVF Conf.
  Computer Vision and Pattern Recognition. pp. 636--645 (2018)

\bibitem{wei2025integral}
Wei, W., Qian, Y., Chen, H., Dai, J., Jin, Y.: Integral fast fourier color
  constancy. In: Proceedings of the Computer Vision and Pattern Recognition
  Conference. pp. 26420--26429 (2025)

\bibitem{woo2017improving}
Woo, S.M., Lee, S.H., Yoo, J.S., Kim, J.O.: Improving color constancy in an
  ambient light environment using the phong reflection model. IEEE Transactions
  on Image Processing  \textbf{27}(4),  1862--1877 (2017)

\bibitem{Yang2015Efficient}
Yang, K.F., Gao, S.B., Li, Y.J.: Efficient illuminant estimation for color
  constancy using grey pixels. In: IEEE/CVF Conf. Computer Vision and Pattern
  Recognition. pp. 2254--2263 (2015)

\bibitem{yu2020cascading}
Yu, H., Chen, K., Wang, K., Qian, Y., Zhang, Z., Jia, K.: Cascading
  convolutional color constancy. In: AAAI Conference on Artificial
  Intelligence. vol.~34, pp. 12725--12732 (2020)

\bibitem{Wei2023ColorCF}
Yue, S., Wei, M.: Color constancy from a pure color view. J. Opt. Soc. Am. A
  \textbf{40}(3),  602--610 (2023)

\bibitem{zhang2023mm}
Zhang, D., Zhou, F., Jiang, Y., Fu, Z.: Mm-bsn: Self-supervised image denoising
  for real-world with multi-mask based on blind-spot network. In: 2023 IEEE/CVF
  Conference on Computer Vision and Pattern Recognition Workshops (CVPRW). pp.
  4189--4198. IEEE (2023)

\bibitem{zhang2016automatic}
Zhang, L., Zhou, H., Yan, L., Zheng, R., Yu, F.: An automatic white balance
  method based on gray world and coincidence of chromaticity histogram. In: 8th
  International Symposium on Advanced Optical Manufacturing and Testing
  Technologies: Optoelectronic Materials and Devices. vol.~9686, pp. 266--273.
  SPIE (2016)

\bibitem{zhang2022idr}
Zhang, Y., Li, D., Law, K.L., Wang, X., Qin, H., Li, H.: Idr: Self-supervised
  image denoising via iterative data refinement. In: 2022 IEEE/CVF Conference
  on Computer Vision and Pattern Recognition (CVPR). pp. 2088--2097. IEEE
  (2022)

\bibitem{zhao2022improving}
Zhao, Z., Hu, H.M., Zhang, H., Chen, F., Guo, Q.: Improving color constancy
  using chromaticity-line prior. IEEE Transactions on Multimedia  \textbf{25},
  3642--3656 (2022)

\end{thebibliography}
